\documentclass{article}
\usepackage{spconf,amsmath,graphicx,hyperref}
\usepackage{booktabs}
\usepackage{comment}

\usepackage{array}
\usepackage{tabularx}

\usepackage{hyperref}
\usepackage{tikz}

\makeatletter
\def\name#1{\gdef\@name{{\em #1}\par}}  
\def\@maketitle{\newpage
  \null
  \vskip 2em \begin{center}
  {\large \bf \@title \par} \vskip 1.5em 
  {\large \lineskip .5em
    \@name 
    \@address
  \par} \end{center}
  \par
  \vskip 1.5em}
\makeatother

\title{Easy Turn: Integrating Acoustic and Linguistic Modalities for Robust Turn-Taking in Full-Duplex Spoken Dialogue Systems}
\name{
    Guojian Li$^{1}$, Chengyou Wang$^{1}$, Hongfei Xue$^{1}$, Shuiyuan Wang$^{1}$, Dehui Gao$^{1}$, Zihan Zhang$^{2}$, Yuke Lin$^{2}$, \\
    Wenjie Li$^{2}$, Longshuai Xiao$^{2}$, Zhonghua Fu$^1$$~^*$, Lei Xie$^1$$~^*$
}
\address{
    $^{1}$Audio, Speech and Language Processing Group (ASLP@NPU), School of Computer Science, 
        \\ Northwestern Polytechnical University, Xi’an, China \\
    $^{2}$Huawei Technologies, China \\
    \texttt{aslp\_lgj@mail.nwpu.edu.cn, mailfzh@nwpu.edu.cn, lxie@nwpu.edu.cn}
}

\begin{document}
\ninept
\maketitle
\let\oldthefootnote\thefootnote               
\renewcommand{\thefootnote}{*}                
\footnotetext{Corresponding authors.}
\renewcommand{\thefootnote}{\oldthefootnote}  
\begin{abstract}
 Full-duplex interaction is crucial for natural human–machine communication, yet remains challenging as it requires robust turn-taking detection to decide when the system should speak, listen, or remain silent. Existing solutions either rely on dedicated turn-taking models, most of which are not open-sourced. The few available ones are limited by their large parameter size or by supporting only a single modality, such as acoustic or linguistic. Alternatively, some approaches finetune LLM backbones to enable full-duplex capability, but this requires large amounts of full-duplex data, which remain scarce in open-source form. To address these issues, we propose \emph{Easy Turn}—an open-source, modular turn-taking detection model that integrates acoustic and linguistic bimodal information to predict four dialogue turn states: \emph{complete}, \emph{incomplete}, \emph{backchannel}, and \emph{wait}, accompanied by the release of \emph{Easy Turn trainset}, a 1,145-hour speech dataset designed for training turn-taking detection models. Compared to existing open-source models like TEN Turn Detection and Smart Turn V2, our model achieves state-of-the-art turn-taking detection accuracy on our open-source \emph{Easy Turn testset}. The data and model will be made publicly available on GitHub.\footnote{https://github.com/ASLP-lab/Easy-Turn}
\end{abstract}
\section{Introduction}
\label{sec:intro}

In human-machine speech interaction, full-duplex interaction enables simultaneous bidirectional information flow, allowing participants to switch smoothly and naturally between speaking and listening roles~\cite{lin2022duplex}. This capability is critical for achieving real-time, efficient interaction~\cite{lu2025duplexmamba}. While full-duplex exchange occurs effortlessly for humans, it remains a major challenge for spoken dialogue systems~\cite{liu2020towards}. To achieve equally natural communication, such systems require robust turn-taking detection, which involves accurately identifying user intent to determine when the system should speak, listen, or remain silent.

With the rapid advancement of LLMs, spoken dialogue systems have made remarkable progress in recent years~\cite{xu2025qwen2, yu2024salmonn, fu2025vita, defossez2024moshi}, quickly entering daily life and exerting a profound impact. The turn-based (half-duplex) approaches already deliver excellent interactive experiences. Research on full-duplex spoken dialogue systems has become a major focus, with current efforts falling into two main categories: 1) integrating an additional turn-taking detection model independent of the LLM backbone, such as TEN Turn Detection\footnote{https://github.com/ten-framework/ten-turn-detection} and Smart Turn V2\footnote{https://github.com/pipecat-ai/smart-turn}. TEN Turn Detection utilizes the Qwen2.5-7B LLM to perform semantic analysis on text inputs, predicting one of three states: \emph{finished}, \emph{unfinished}, or \emph{wait}. Smart Turn V2 is the latest turn-taking detection model of the Smart Turn series. It employs Wav2Vec2~\cite{baevski2020wav2vec} as the acoustic encoder with a linear classifier layer to directly conduct turn-taking detection from speech, outputting one of two states: \emph{complete} or \emph{incomplete}. 2) endowing the LLM backbone with full-duplex capability through fine-tuning on large-scale full-duplex dialogue data, as seen in Moshi~\cite{defossez2024moshi}, Freeze-Omni~\cite{wang2024freeze}, and GLM-4-Voice~\cite{zeng2024glm}, etc.


Although full-duplex spoken dialogue systems have achieved notable progress, they continue to face substantial challenges. For the approach that integrates a standalone turn-taking detection model, most implementations remain closed-source. Only a few available open-source models include TEN Turn Detection and Smart Turn V2. However, both models have significant limitations: 1) TEN Turn Detection suffers from an excessively large 7B parameter size and lacks direct speech input support, requiring an upstream automatic speech recognition (ASR) model that risks losing critical acoustic information. 2) Smart Turn V2 is fast and lightweight but relies solely on a linear classifier layer, making it difficult to capture rich semantic information. For the approach that directly trains the LLM backbone to acquire full-duplex capabilities, it is even more demanding, requiring extensive high-quality full-duplex dialogue data. Yet, most studies either withhold full-duplex training datasets entirely or provide only a brief description of their processing pipelines~\cite{zhang2025llm}.

To accelerate research in this field, we introduce \emph{Easy Turn}, an open-source and modular turn-taking detection model. The model accepts user's speech as input and outputs both the corresponding ASR transcription and the dialogue turn state, effectively integrating acoustic and linguistic information. More importantly, we concurrently release \emph{Easy Turn trainset}, a large-scale speech dataset specifically designed for training turn-taking detection models. This dataset spans all four dialogue turn states (\emph{complete}, \emph{incomplete}, \emph{backchannel}, \emph{wait}) with a total duration of 1,145 hours. Compared with the existing open-source turn-taking detection models TEN Turn Detection and Smart Turn V2, Easy Turn supports the most comprehensive set of dialogue turn state categories: \emph{complete} (semantically complete), \emph{incomplete} (semantically incomplete), \emph{backchannel} (brief feedback), and \emph{wait} (request to pause or end the dialogue). On our open-source \emph{Easy Turn testset}, Easy Turn achieves state-of-the-art state classification accuracy while maintaining competitive inference speed and memory efficiency. The data and model will be made publicly available on GitHub.

\begin{figure}[t]
    \centering
    \includegraphics[width=1.0\linewidth]{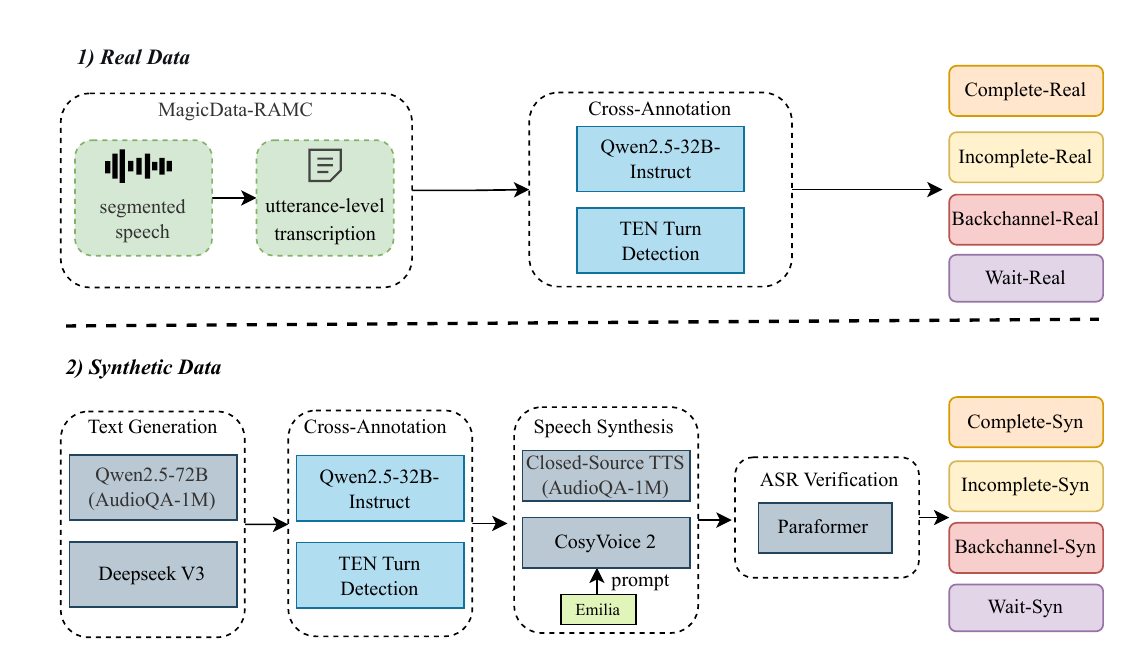}
    \caption{The detailed processing pipeline of the Easy Turn Corpus.}
    \label{fig:pipeline}
\end{figure} 

\section{Easy Turn Corpus}
\label{sec:format}
Our Easy Turn corpus consists of two parts: the Easy Turn trainset and the Easy Turn testset, covering four dialogue turn states: \emph{complete}, \emph{incomplete}, \emph{backchannel}, and \emph{wait}. Each speech recording is paired with a transcription and labeled with one of these states. In the following sections, we detail the definition of each turn state and the construction process of both the trainset and testset.
\subsection{Turn States}

\hspace*{1.5em}\textbf{Complete} The complete state indicates that the user has fully expressed their intent and expects an immediate response from the spoken dialogue system, as in the utterance, “Can you tell me a story?”. A high-performance full-duplex spoken dialogue system should respond swiftly and accurately to such complete expressions. 

\textbf{Incomplete} The incomplete state occurs when a user pauses but clearly has not finished speaking, as in “I want to ask…” or “Do you know how it…”. An ideal full-duplex spoken dialogue system continues listening until the user’s semantic expression is complete, rather than interrupting prematurely. 

\textbf{Backchannel} In natural conversations, backchannel signals refer to brief listener responses (e.g., “Uh-huh.”, “Right.”) that indicate active engagement and comprehension while the speaker is talking~\cite{lin2025full}. In full-duplex spoken dialogue systems, backchannel signals from users should not interrupt the system’s speech output, which is critical for maintaining interaction fluency and enhancing user experience. 

\textbf{Wait} The wait state refers to cases where users explicitly request to pause or terminate the interaction with the spoken dialogue system (e.g., “shut up”, “please stop”), serving as an efficient and concise way to end the system’s current turn or completely halt the dialogue~\cite{xie2024mini}.

\subsection{Trainset}
The Easy Turn trainset is a large-scale speech dataset designed for training turn-taking detection models, comprising both real and synthetic data. It comprises four dialogue turn states (\emph{complete}, \emph{incomplete}, \emph{backchannel}, \emph{wait}), totaling approximately 1,145 hours. Detailed statistics are provided in Table~\ref{tab:trainset}. The detailed data processing pipeline is shown in Figure~\ref{fig:pipeline}. The following sections detail the procedures used to construct them.

\begin{table}[htbp]
\caption{The statistics of the Easy Turn trainset.}
\label{tab:trainset}
\centering
\scalebox{0.8}{ 
\begin{tabular}{lcc}
\toprule
\textbf{State} & \textbf{Entries (k)} & \textbf{Speech Duration (h)} \\
\midrule
Complete & 423 & 580 \\
Incomplete & 712  & 532 \\
Backchannel & 41 & 10 \\
Wait & 40 & 23 \\
\bottomrule
\end{tabular}
}
\end{table}

\subsubsection{Real Data}
We derive four real-scenario subsets from the MagicData-RAMC corpus~\cite{yang2022open}, one of the largest publicly available Mandarin dialogue datasets, which contains approximately 180 hours of spontaneous speech with transcriptions manually labeled and verified by professional annotators. For all subsets, we first segment long conversations into utterance-level samples using the timestamps provided in the dataset and align them with the corresponding transcriptions. Since no reliable and efficient open-source speech annotation tools exist for turn-taking detection, we design prompt templates to automate text labeling with text-based LLMs, targeting four dialogue turn states: \emph{complete}, \emph{incomplete}, \emph{backchannel}, and \emph{wait}.

\textbf{Complete-Real} We apply a cross-annotation strategy that retains only utterances labeled as \emph{complete} by both Qwen2.5-32B-Instruct~\cite{an2024qwen2.5}, a versatile instruction-tuned LLM, and TEN Turn Detection, our previously introduced turn-taking detection model. This subset represents semantically complete cases in real conversational scenarios.

\textbf{Incomplete-Real} Following the same strategy, we retain only utterances labeled as \emph{incomplete} by both models, forming the Incomplete-Real subset, which represents semantically incomplete cases in real conversational scenarios.

\textbf{Backchannel-Real} For this subset, we focus on speech samples shorter than 2 seconds. Because TEN Turn Detection does not support backchannel labeling, we rely solely on Qwen2.5-32B-Instruct to annotate the transcriptions, keeping only those labeled as \emph{backchannel}. This subset reflects brief listener feedback signals that naturally occur in real conversation.

\textbf{Wait-Real} Following the same strategy as for the Complete-Real subset, we retain only utterances labeled as \emph{wait} by both models, thereby forming the Wait-Real subset, which captures real conversational cases where a speaker requests to pause or terminate the dialogue.

\subsubsection{Synthetic Data}
 To compensate for the lack of human–machine interaction scenarios in the real subsets, we construct four synthetic subsets: Complete-Syn, Incomplete-Syn, Backchannel-Syn, and Wait-Syn—through a combination of large-scale text generation, cross-annotation, speech synthesis, and ASR verification. For all subsets, we generate diverse text samples using either DeepSeek V3~\cite{liu2024deepseek} or Qwen2.5-72B, apply the same cross-annotation strategy as in the real data to filter texts, synthesize speech with CosyVoice 2~\cite{du2024cosyvoice} and others, with the Emilia corpus~\cite{he2025emilia} serving as the reference speech source. Finally, we validate the synthetic outputs with Paraformer~\cite{gao2022paraformer}, retaining only samples with zero word error rate (WER) to ensure high quality.
 
\textbf{Complete-Syn} We start from the AudioQA-1M corpus~\cite{gao2025lucy}, a large-scale corpus where Qwen2.5-72B generates question–answer dialogues subsequently converted to speech via closed-source text-to-speech (TTS) system. We retain only the question segments to better simulate real spoken queries. Following the same cross-annotation strategy as in Complete-Real subset, we apply Qwen2.5-32B-Instruct and TEN Turn Detection to the transcriptions of question segments, yielding the Complete-Syn subset.

\textbf{Incomplete-Syn} We generate a large collection of semantically incomplete texts with Deepseek V3, then apply cross-annotation with Qwen2.5-32B-Instruct and TEN Turn Detection, keeping only those labeled as \emph{incomplete}. During speech synthesis, we deliberately insert elongated endings or 0–1 second pauses to simulate natural hesitations or thought-induced interruptions, thereby enriching the realism of incomplete utterances.

\textbf{Backchannel-Syn} To address the limited diversity of user backchannel expressions in Backchannel-Real, we use Deepseek V3 to generate a broader range of backchannel texts. These are annotated by Qwen2.5-32B-Instruct, retaining only samples labeled as \emph{backchannel}, before undergoing the standard synthesis and validation process.

\textbf{Wait-Syn} We generate a diverse text set of pause or termination requests using Deepseek V3. After cross-annotation with both LLMs, we retain only utterances labeled as \emph{wait}, then synthesize and validate them using the same pipeline, forming the Wait-Syn subset.

\subsection{Testset}

In addition to the Easy Turn trainset, we also release a speech test set—Easy Turn testset, designed to evaluate turn-taking detection performance. It includes four dialogue turn states: 300 samples each for complete and incomplete, and 100 samples each for backchannel and wait. Real and synthetic samples are balanced at a 1:1 ratio. The transcriptions of Easy Turn testset come from sources outside the trainset, covering both casual conversations and human-machine interactions. Dialogue turn states are manually annotated to ensure higher accuracy. The Easy Turn testset includes two types of speech: real recordings from human speakers and synthetic speech generated with CosyVoice 2. These designs ensure the independence and diversity of the test set.

\begin{table*}[htbp]
\caption{Performance comparison of different approaches. The symbol “–” indicates that the corresponding model does not support detection of a particular state. 
}
\label{tab:different}
\centering
\scalebox{0.88}{ 
\begin{tabular}{lccccccc}
\toprule
 \textbf{Model} & \textbf{Params(MB)} $\downarrow$& \textbf{Latency(ms)} & 
 \textbf{Memory(MB)} & \textbf{{$\text{ACC}_{cp}$}(\%)} $\uparrow$& \textbf{{$\text{ACC}_{incp}$}(\%)}  & \textbf{{$\text{ACC}_{bc}$}(\%)} & \textbf{{$\text{ACC}_{wait}$}(\%)}  \\
\midrule
Paraformer + TEN Turn Detection & 7220 & 204 & 15419 & 86.67 & 89.3 & - & 91  \\
 Smart Turn V2 & \textbf{95} & \textbf{27} & \textbf{370} & 78.67 & 62 & - & -   \\
 Easy Turn (Proposed)  & 850 & 263 & 2559 & \textbf{96.33}  & \textbf{97.67} & \textbf{91} & \textbf{98} \\
\bottomrule
\end{tabular}
}
\end{table*}

\section{Easy Turn}
\subsection{Model Architecture}
\begin{figure}[htbp]
    \centering
    \includegraphics[width=0.95\columnwidth]{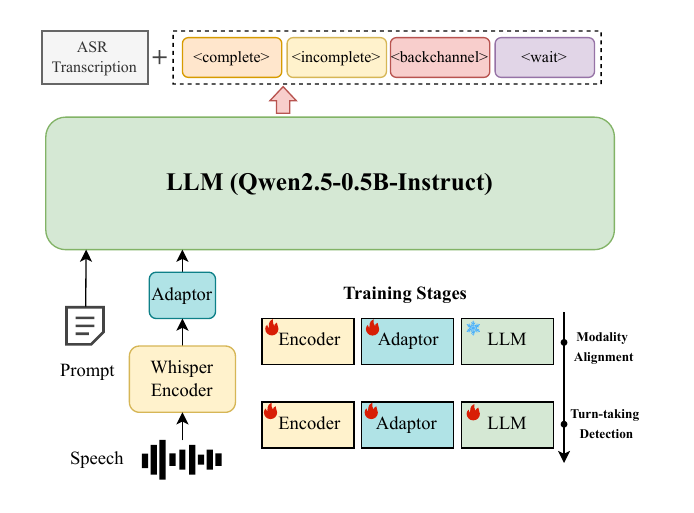}
    \caption{The architecture of the proposed model and training stages.}
    \label{fig:architecture}
\end{figure}

Our turn-taking detection model, Easy Turn, comprises three key components: an audio encoder, an audio adaptor, and an LLM. Its design draws inspiration from Qwen-Audio~\cite{chu2023qwen}, employing Whisper~\cite{radford2023robust} as the audio encoder and Qwen2.5 as the LLM. The overall architecture of Easy Turn is illustrated in Figure~\ref{fig:architecture}.

Specifically, we adopt the Whisper-Medium model as the audio encoder, which includes two 1D convolutional layers and 24 Transformer layers. This configuration achieves an effective balance between speech comprehension capability and inference efficiency. The adaptor module employs a hybrid architecture comprising three 1D convolutional layers and four Transformer layers to efficiently bridge the output of the audio encoder with the input requirements of the LLM, enabling robust modality alignment~\cite{xu2025adaptor}. Unlike Qwen-Audio, which leverages a large-scale LLM for multitask processing, we focus solely on turn-taking detection and select the more lightweight Qwen2.5-0.5B-Instruct~\cite{an2024qwen2.5}, ensuring higher training and inference efficiency.

\subsection{ASR+Turn-Detection}
To better integrate acoustic and linguistic modalities, we adopt an ASR+Turn-Detection paradigm~\cite{geng2025osum}, where the LLM generates ASR transcriptions and fuses them with acoustic features to sequentially predict dialogue turn state labels (\emph{complete}, \emph{incomplete}, \emph{backchannel} or \emph{wait}). To implement this paradigm, we design natural language prompts to guide the LLM. Deepseek V3~\cite{liu2024deepseek} generates five candidate prompts, with one randomly sampled during training to enhance robustness. Easy Turn takes speech signals and natural language prompts as input and produces corresponding ASR transcriptions and dialogue turn state labels following the ASR+Turn-Detection paradigm. 

\section{EXPERIMENTS}
\label{sec:pagestyle}

\subsection{Experimental Setup}

Our model training consists of two stages: modality alignment training and dedicated training for dialogue turn-taking detection task.

In the first stage, modality alignment training aims to align the acoustic modality representation with the linguistic modality representation. We conduct alignment training on the ASR task with 23,000 hours of data, including the Aishell1~\cite{bu2017aishell}, Aishell2~\cite{du2018aishell}, and WenetSpeech~\cite{zhang2022wenetspeech} open-source corpus, as well as carefully annotated internal ASR data. During this stage, all LLM parameters remain frozen, while only the audio encoder and adaptor are updated. Training uses a learning rate of 5e-5 and a batch size of 16, running for 3 epochs on 8 NVIDIA RTX 4090 GPUs with the WeNet toolkit~\cite{yao2021wenet}. In the second stage, the model undergoes dedicated training for the dialogue turn-taking detection task using the previously constructed Easy Turn trainset. In this stage, we unfreeze the audio encoder, adaptor, and LLM parameters for full fine-tuning to maximize model performance. Training uses a learning rate of 2e-5 and a batch size of 12, running for 6 epochs on 8 NVIDIA RTX 4090 GPUs, also implemented with the WeNet toolkit.

During inference, the model runs on a single NVIDIA RTX 4090 GPU. Sampling is disabled, temperature is fixed at 1.0, and greedy search is employed to ensure generation accuracy while maintaining inference efficiency.

\subsection{Main Results}
We evaluate Easy Turn against two open-source turn-taking detection models, TEN Turn Detection and Smart Turn V2, using the Easy Turn testset. All experiments are conducted on a single NVIDIA RTX 4090 GPU. Notably, since TEN Turn Detection lacks direct speech support, We use Paraformer as the ASR model to transcribe speech into text and take the text as its input. Table~\ref{tab:different} reports the results: $\text{ACC}_{cp}$, $\text{ACC}_{incp}$, $\text{ACC}_{bc}$ and $\text{ACC}_{wait}$ denote the turn-taking detection accuracy for \emph{complete}, \emph{incomplete}, \emph{backchannel}, and \emph{wait} states (higher is better); Params, Latency, and Memory represent total model size, average inference time, and GPU usage, where lower values indicate greater efficiency.

As shown in Table~\ref{tab:different}, Easy Turn supports the most comprehensive set of dialogue turn states and achieves the highest accuracy across all categories. Its ASR + Turn-Detection paradigm effectively integrates acoustic and linguistic modalities, yielding strong performance with only a slight increase in latency, while keeping parameters and memory usage low for easy deployment. In comparison, Ten Turn Detection delivers good accuracy but requires a 7B-parameter LLM and an additional ASR module (Paraformer~\cite{gao2022paraformer}), leading to heavy resource demands. Smart Turn V2 is efficient in latency and memory but suffers from low accuracy and supports only two dialogue turn states, limiting its practicality.

\subsection{Ablation Study}
We conduct ablation experiments on the Easy Turn to evaluate the contribution of individual modalities within its architecture and to assess the impact of the ASR + Turn-Detection paradigm on performance. The primary metric is $\text{ACC}_\text{avg}$, computed as the average detection accuracy across four dialogue turn states (\emph{complete}, \emph{incomplete}, \emph{backchannel}, and \emph{wait}). \text{Easy Turn}\textsubscript{only-state} uses the same architecture as Easy Turn but omits the ASR + Turn-Detection paradigm, directly predicting dialogue turn state labels without first generating ASR transcriptions. Finetuned Whisper + Linear refers to fine-tuning Whisper-Medium audio encoder~\cite{radford2023robust} with an additional linear classifier on our Easy Turn trainset, taking only speech as input and directly predicting dialogue turn state labels, representing the acoustic-only modality. Finetuned Qwen2.5-0.5B-Instruct~\cite{an2024qwen2.5} is fine-tuned on text transcriptions from our Easy Turn trainset, taking only text as input and outputting dialogue turn state labels, representing the linguistic-only modality. Detailed results are shown in Table~\ref{tab:ablation}. 

\begin{table}[htbp]
\caption{Performance comparison of ablation components.}
\label{tab:ablation}
\centering
\scalebox{0.9}{ 
\begin{tabular}{lcc}
\toprule
\textbf{Model} & \textbf{Modality} & \textbf{$\text{ACC}_\text{avg}$ $\uparrow$} \\
\midrule
Easy Turn (Proposed)  & Acoustic+Linguistic  & \textbf{95.75} \\
Easy Turn\textsubscript{only-state} & Acoustic+Linguistic  & 87.88 \\
Finetuned Qwen2.5-0.5B-Instruct & Linguistic-only & 86.25
\\
Finetuned Whisper + Linear & Acoustic-only & 85.50 \\
\bottomrule
\end{tabular}
}
\end{table}

As Table~\ref{tab:ablation} demonstrates, both acoustic-only and linguistic-only models perform poorly, and excluding the ASR + Turn-detection paradigm significantly degrades performance. In contrast, Easy Turn integrates a Whisper-Medium audio encoder, an audio adaptor, and a lightweight 0.5B-parameter LLM, while adopting the ASR + Turn-detection paradigm to effectively fuse acoustic and linguistic modalities, thereby achieving superior performance.

\begin{figure}[htbp]
    \centering
    \includegraphics[width=1.05\columnwidth]{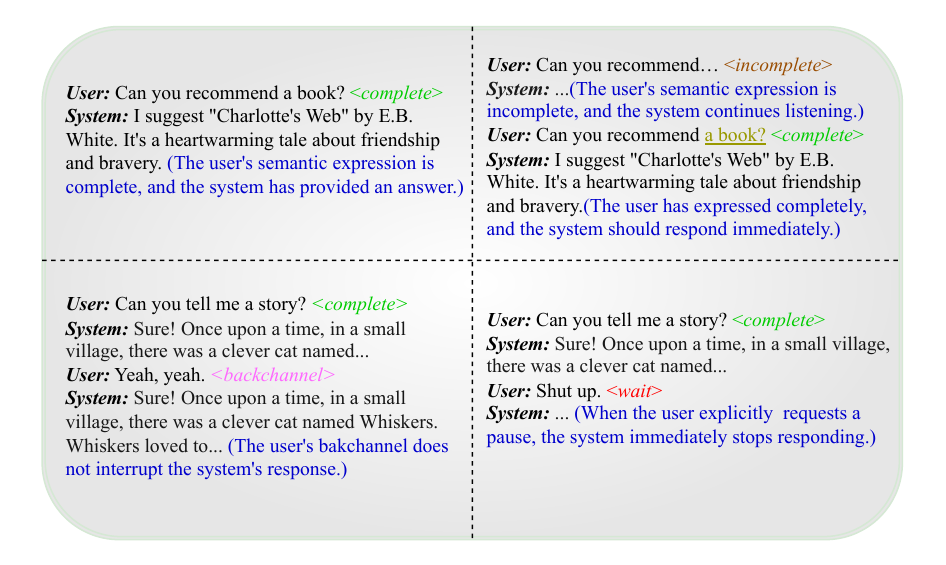}
    \caption{Some examples of Easy Turn applications in spoken dialogue systems.}
    \label{fig:examples}
\end{figure} 

\subsection{Examples}
In Figure~\ref{fig:examples}, we present several examples of Easy Turn applications in spoken dialogue systems. The content inside the angle brackets indicates the dialogue turn state detected by Easy Turn, while the text in parentheses represents the actions the system should take based on the detected dialogue turn state. To evaluate its performance in turn-taking detection, we deploy Easy Turn in our internal spoken dialogue system, where human users interact with the system through microphone input. The results show that Easy Turn performs effectively, accurately identifying dialogue turn states and enabling the system to respond appropriately.

\section{CONCLUSION}
\label{sec:majhead}
We present Easy Turn, a turn-taking detection model for full-duplex spoken dialogue systems. It integrates a Whisper-Medium audio encoder, an audio adaptor, and a lightweight 0.5B-parameter LLM, adopting the ASR + Turn-Detection paradigm to fuse acoustic and linguistic modalities. On the Easy Turn testset, Easy Turn achieves state-of-the-art performance over existing open-source models. We also release the Easy Turn trainset with a detailed construction process, providing a valuable resource for full-duplex dialogue research. Future work will scale training data and explore more diverse dialogue scenarios to improve robustness and generalization.




\begin{small}
\bibliographystyle{IEEEbib}
\bibliography{strings,refs}
\end{small}
\end{document}